%% file: Template.tex
\documentclass{article}
\usepackage{spconf,amsmath,graphicx,hyperref}
\usepackage[utf8]{inputenc} 
\usepackage[T1]{fontenc}    
\usepackage{hyperref}       
\usepackage{url}            
\usepackage{booktabs}       
\usepackage{amsfonts}       
\usepackage{amsmath, bm}
\usepackage{mathrsfs} 
\usepackage{amsthm}
\usepackage{nicefrac}       
\usepackage{microtype}      
\usepackage{xcolor}         

\usepackage{graphicx}
\usepackage{caption}
\usepackage{subcaption}
\usepackage{enumerate}
\usepackage{cite}

\usepackage{tikz}
\usepackage{graphics}

\usepackage{amsfonts}
\usepackage{amsmath}
\usepackage{amssymb}
\usepackage{mathtools}
\usepackage{amsthm}
\usepackage{bbm}

\usepackage[capitalize,noabbrev]{cleveref}

\usepackage{algorithm}
\usepackage{algpseudocode}

\usepackage{enumitem}

\input{mysymb}

\title{Unrolled Graph Neural Networks for Constrained Optimization}
\name{Samar Hadou \qquad Alejandro Ribeiro
}
\address{Department of Electrical and Systems Engineering, University of Pennsylvania, PA, USA}
\begin{document}
\ninept
\maketitle
\input{sections/abstract}
\begin{keywords}
Unrolling, constrained optimization, GNNs, dual ascent, constrained learning
\end{keywords}

\input{sections/intro}

\input{sections/architecture}
\input{sections/training}

\input{sections/numerical}

\input{sections/conclusions}

\bibliographystyle{IEEEbib}
\bibliography{Bib}

\end{document}

%% file: mysymb.tex
\newcounter{excercise}
\newcounter{excercisepart}

\def \reals    {{\mathbb R}}

\def\mbE{{\ensuremath{\mathbb E}}}

\def\ccalC{{\ensuremath{\mathcal C}}}
\def\ccalD{{\ensuremath{\mathcal D}}}

\def\ccalI{{\ensuremath{\mathcal I}}}

\def\ccalL{{\ensuremath{\mathcal L}}}

\def\ccal0{{\ensuremath{\mathcal 0}}}
\def\bbA{{\ensuremath{\mathbf A}}}

\def\bbM{{\ensuremath{\mathbf M}}}

\def\bbP{{\ensuremath{\mathbf P}}}

\def\bbW{{\ensuremath{\mathbf W}}}

\def\bbS{{\ensuremath{\mathbf S}}}

\def\bbX{{\ensuremath{\mathbf X}}}

\def\bbb{{\ensuremath{\mathbf b}}}
\def\bbc{{\ensuremath{\mathbf c}}}

\def\bbe{{\ensuremath{\mathbf e}}}
\def\bbf{{\ensuremath{\mathbf f}}}

\def\bbq{{\ensuremath{\mathbf q}}}

\def\bbx{{\ensuremath{\mathbf x}}}
\def\bby{{\ensuremath{\mathbf y}}}
\def\bbz{{\ensuremath{\mathbf z}}}
\def\bb0{{\ensuremath{\mathbf 0}}}
\def\rmD{{\ensuremath\text{D}}}

\def\rmP{{\ensuremath\text{P}}}

\def\rmh{{\ensuremath\text{h}}}

\def\bbtheta{\boldsymbol{\theta}}

\def\bblambda{\boldsymbol{\lambda}}

\def\bbmu{\boldsymbol{\mu}}
\def\bbnu{\boldsymbol{\nu}}

\def\bbTheta{\boldsymbol{\Theta}}

\def\bbPhi{\boldsymbol{\Phi}}

\DeclarePairedDelimiterX{\infdivx}[2]{(}{)}{%
  #1\;\delimsize\|\;#2%
}

\newcommand{\pr}[1]{\mathbb{#1}}

\makeatletter
\DeclareRobustCommand{\cev}[1]{%
  {\mathpalette\do@cev{#1}}%
}
\newcommand{\do@cev}[2]{%
  \vbox{\offinterlineskip
    \sbox\z@{$\m@th#1 x$}%
    \ialign{##\cr
      \hidewidth\reflectbox{$\m@th#1\vec{}\mkern4mu$}\hidewidth\cr
      \noalign{\kern-\ht\z@}
      $\m@th#1#2$\cr
    }%
  }%
}

\makeatother

%% file: sections/abstract.tex
\begin{abstract}
    In this paper, we unroll the dynamics of the dual ascent (DA) algorithm in two coupled graph neural networks (GNNs) to solve constrained optimization problems. The two networks interact with each other at the layer level to find a saddle point of the Lagrangian. The primal GNN finds a stationary point for a given dual multiplier, while the dual network iteratively refines its estimates to reach an optimal solution. We force the primal and dual networks to mirror the dynamics of the DA algorithm by imposing descent and ascent constraints. We propose a joint training scheme that alternates between updating the primal and dual networks. Our numerical experiments demonstrate that our approach yields near-optimal near-feasible solutions and generalizes well to out-of-distribution (OOD) problems.
\end{abstract}

%% file: sections/intro.tex
\section{Introduction}

Algorithm unrolling has emerged as a learning-to-optimize paradigm that embeds iterative solvers into neural network architectures, yielding learnable optimizers endowed with domain-specific knowledge \cite{monga_algorithm_2021,gregor_learning_2010}.
Despite the extensive literature on unrolling, most work focuses on unconstrained optimization problems, e.g., \cite{gregor_learning_2010,zhang2020deep, mou2022deep, Savanier2023Deep, wang2025proximal,hadou2024Robust,wang2024ufc, zhou2024deep, fang2024real, he2025run}. 
For constrained settings, a common approach is to incorporate constraints into the training loss as penalty terms \cite{kishida2022temporal,bertocchi2020deep,hadou2023stochastic}. Another line of work considers unrolling primal-dual algorithms \cite{nagahama2022graph, Zhang2025, Yang2024, li2024pdhg}, but these efforts typically restrict learning to a few scalar hyperparameters, such as the step size, rather than learning the full algorithmic dynamics. As a result, such approaches remain narrowly tailored to the designated  problem classes.

In this paper, we unroll the dynamics of dual ascent (DA) into two coupled graph neural networks (GNNs). The models aim to collaboratively find the saddle point of the Lagrangian function associated with a constrained problem. The primal GNN predicts a layerwise trajectory towards a stationary point of the Lagrangian for a given multiplier, while the dual GNN uses the primal predictions to ascend the dual function toward its optimum. 
We cast the training problem as a nested optimization problem, where the inner level trains the primal network over a joint distribution of constrained problems and Lagrangian multipliers, while the outer level trains the dual network. Crucially, the multiplier distribution required by the primal network training is not known \emph{a priori}, as it is induced by the multiplier trajectories generated by the forward pass of the dual network. Our nested formulation allows for alternating training steps to overcome this challenge. The dual network first generates multiplier trajectories to train the primal network, whose solutions are then used in the dual network updates. This procedure optimizes each network against the most recent outputs of the other until convergence.

Departing from traditional unrolling literature, which wires a specific iterative algorithm into the architecture, we instead use standard GNNs to model the primal and dual iterates to leverage their expressivity and permutation equivariance. GNNs are well suited to constrained problems, as variable-constraint relations can be naturally encoded in a graph adjacency matrix \cite{cappart2023combinatorial, gasse2019exact,chenrepresenting}.  However, this design choice can further exacerbate the lack of monotonic descent often observed in unrolled networks.
As reported in \cite{hadou2024Robust}, the lack of descent behavior makes the networks prone to perturbations and distribution shifts.
To address this limitation, we impose monotonic descent and ascent constraints on the outputs of the primal and dual layers, respectively, during training. These constraints encourage the models to learn the underlying DA dynamics rather than merely fitting the final solution, which improves the final-layer performance and enhances out-of-distribution (OOD) generalization, as demonstrated in our numerical experiments.

In summary, the main contributions of this work are as follows.
\begin{itemize}
    \item We design a pair of GNNs that interact at the layer level to solve the dual problem (Section~\ref{sec:architecture}).
    \item We formulate the training problem as a nested optimization problem and impose descent and ascent constraints on the unrolled layers (Section~\ref{sec:training}).
    \item We empirically assess our approach on mixed-integer quadratic programs (MIQPs) with linear constraints (Section~\ref{sec:simulations})
\end{itemize}

%% file: sections/architecture.tex
\section{Constrained Optimization}

Consider a constrained problem that poses the task of minimizing a scalar objective function $f_0: \mathbb{R}^n \to \mathbb{R}$ subject to $m$ constraints, 
\begin{equation} \label{eq:problem}
    \begin{split}
        P^*(\bbz) ~=~ \underset{\bbx \in \mathbb{R}^n}{\text{min}} \quad  f_0(\bbx;\bbz) \quad
        \text{s.t.}\quad  {\bf f}(\bbx; \bbz) \leq {\bf 0},
    \end{split}
\end{equation}
where ${\bf f}: \mathbb{R}^n \to \mathbb{R}^m$ is a vector-valued function representing the problem constraints, and $\bbz$ represents a \emph{problem instance}. Moving to the dual domain, the Lagrangian function, ${\cal L}: \mathbb{R}^n \times \mathbb{R}^m_+ \to \mathbb{R}$, associated with \eqref{eq:problem} is
\begin{equation}
    {\ccalL}(\bbx, \boldsymbol{\lambda}; \bbz) ~=~ f_0(\bbx;{\bbz}) + \boldsymbol{\lambda}^\top {\bf f}(\bbx;{\bbz}),
\end{equation}
where $\boldsymbol{\lambda}$ contains the dual multipliers. The dual problem is defined as 
\begin{align}\label{eq:saddle-point}
    D^*(\bbz) ~=~ \underset{\bblambda \in \pr{R}^m_+}{\text{max}} \, 
                \underset{\bbx}{\min}~~{\ccalL}(\bbx, \boldsymbol{\lambda}; \bbz),
\end{align}
and the duality theory affirms that $D^*(\bbz) \leq P^*(\bbz)$ \cite{boyd2004convex}. The equality holds under strong duality, which applies to convex problems. 
Under this assumption, the Lagrangian has a saddle point $(\bbx^*, \bblambda^*)$, with $\bbx^*$ and $\bblambda^*$ optimal for \eqref{eq:problem} and \eqref{eq:saddle-point}, respectively. The DA algorithm retrieves the dual optimum $\bblambda^*$ through the iterations:
\vspace{-1pt}
\begin{align}
        \bbx_l^*(\bblambda_l)      &  ~\in~ \underset{\bbx}{\text{argmin}} \ 
                                        {\cal L}\big(\bbx, \bblambda_l; \bbz \big), \nonumber
        \\
        \boldsymbol{\lambda}_{l+1} & ~=~ \Big[ \, \boldsymbol{\lambda}_{l} + \eta 
                                        \ \bbf \big(\bbx_l^*, \bbz \big) \, \Big]_+, \label{eq:DA_updates}    
\end{align}
\vspace{-1pt}
where $\eta$ is a step size, and $[\cdot]_+$ denotes a projection onto $\reals_+^m$. In \eqref{eq:DA_updates}, a Lagrangian stationary point is attained for the recent multiplier $\bblambda_l$, before a (projected) gradient ascent step is taken in the dual domain.
The primal optimum $\bbx^*$ is then recovered from $\bbx^* \in \bbx^*(\bblambda^*)$.

\section{Unrolled Networks for Constrained Optimizations} \label{sec:architecture}

We design a pair of unrolled GNNs that jointly find the saddle point of \eqref{eq:saddle-point}, mimicking the DA dynamics, as illustrated in Fig.~\ref{fig:architecture}. The primal network, $\bbPhi_\rmP$, emulates the minimization problem with respect to $\bbx$, while the dual network, $\bbPhi_\rmD$, emulates the gradient ascent steps in the dual domain across its layers. Each unrolled layer consists of $T$ graph sub-layers, each comprising a graph filter and a nonlinearity \cite{hadou2022space}, followed by a readout that outputs a primal or dual estimate.

The primal network, denoted by $\bbPhi_\rmP(\, \cdot, \cdot\, ;\bbtheta_\rmP)$, predicts a $K$-step trajectory from an initial point $\widetilde\bbx_0$ towards $\widetilde\bbx_K \approx \bbx^*(\bblambda)$ across its $K$ unrolled layers. For a given dual multiplier $\bblambda$ and a problem instance $\bbz$, the $k$-th primal layer refines the estimate $\widetilde\bbx_{k-1}$ into

\begin{equation}\label{eq:primal_unrolling}
      \widetilde\bbx_k ~=~ \bbPhi_\rmP^k \Big(\widetilde\bbx_{k-1}, \bblambda, \bbz;\bbtheta_\rmP^k\Big),
\end{equation}
where $\bbtheta_\rmP^k$ contains the parameters of the graph sub-layers and readout. We use the tilde notation to distinguish the intermediate estimates produced by the primal layers from other occurrences of $\bbx$. 
As a general design principle, we incorporate residual connections between unrolled layers as they naturally mimic gradient-based updates.

The dual network, denoted by $\bbPhi_\rmD(\, \cdot\,;\bbtheta_\rmD, \bbtheta_\rmP)$, has $L$ layers whose outputs constitute a trajectory in the dual domain starting from an initial point $\bblambda_0$ and ending at an estimate of the optimal multiplier, $\bblambda_L \approx \bblambda^*$. 
The $l$-th dual layer is defined as
\begin{equation}\label{eq:dual_layer_unrolling}
    \bblambda_l ~=~ \bbPhi_\rmD^l\Big(\, \bblambda_{l-1}, \, \bbPhi_\rmP(\bblambda_{l-1}, \bbz; \bbtheta_\rmP), \, \bbz; \, \bbtheta_\rmD^l \, \Big),
\end{equation}
where $\bbtheta_\rmD^l$ is the learnable parameters. 
In \eqref{eq:dual_layer_unrolling}, the $l$-th dual layer queries the primal network for its estimate $\bbPhi_\rmP(\bblambda_{l-1}, \bbz; \bbtheta_\rmP)  \approx \bbx^*(\bblambda_{l-1})$, denoted as $\bbx_{l-1}$. Consequently, a single forward pass of the dual network triggers $L$ forward passes of the primal network, as illustrated in Fig.~\ref{fig:architecture}. 
The nonlinearity at the end of each dual layer is chosen as a relu function to ensure that the predicted multipliers are nonnegative.

Finally, the solution to \eqref{eq:problem} is obtained by feeding the final dual estimate, $\bblambda_L = \bbPhi_\rmD(\bbz;\bbtheta_\rmD, \bbtheta_\rmP^*)$, to the primal network,
\begin{equation}
    \bbx_L = \bbPhi_\rmP \Big(\, \bbPhi_\rmD\big(\bbz; \bbtheta_\rmD, \bbtheta_\rmP\big),\,  \bbz;\, 
    \bbtheta_\rmP \, \Big).
\end{equation}

Our goal is to train the primal and dual networks such that the output of the primal network satisfies $\widetilde\bbx_K \approx \bbx^*(\bblambda)$ for any $\bblambda$, and the output of the dual network satisfies $\bblambda_L \approx \bblambda^*$ for a family of optimization problems. In addition, the predicted trajectories should follow descent and ascent dynamics with respect to the corresponding optimization variable.

\input{figures/ActorCritic}

%% file: figures/ActorCritic.tex
\usetikzlibrary{positioning,calc}
\tikzset{
  block/.style={
    draw=black,
    fill=teal,
    fill opacity=0.5,
    text opacity=1.0,
    rectangle,
    minimum width=10em,
    minimum height=2.5em
  },
  dualblock/.style={
    draw=black,
    fill=blue,
    fill opacity=0.4,
    text opacity=1.0,
    rectangle,
    minimum width=10em,
    minimum height=2.5em
  },
  arrow/.style={
    ->,
    >=latex,
    font=\small
  },
  redarrow/.style={
    arrow,
    draw=blue
  },
  tealarrow/.style={
    arrow,
    draw=teal
  },
  vdots/.style={
    font=\Large,
    inner sep=0pt,
    outer sep=0pt
  }
}

\begin{figure}[t]
    \centering
    \resizebox{0.95\columnwidth}{!}{
    \begin{tikzpicture}[node distance=0.8cm and 3cm, auto]
    
      \node[block]               (p1)  {$\Phi_\rmP^{1}\big(\, \widetilde\bbx_0,\bblambda_l, \bbz\,;\theta_\rmP^{1}\big)$};
      \node[block, below=of p1]  (p2)  {$\Phi_\rmP^{2}\big(\,\widetilde\bbx_1, \bblambda_l,\bbz\,;\theta_\rmP^{2}\big)$};
      \node[vdots, below=of p2]  (pd)  {$\vdots$};
      \node[block, below=of pd]  (pL)  {$\Phi_\rmP^{K}\big(\,\widetilde\bbx_{K-1},\bblambda_l,\bbz\,;\theta_\rmP^{K}\big)$};
      \draw[arrow] (p1) -- (p2);
      \draw[arrow] (p2) -- (pd);
      \draw[arrow] (pd) -- (pL);
      \node[above=27pt of p1.north] {$\bbPhi_\rmP(\bblambda_l, \bbz; \bbtheta_\rmP)$};
    
      \node[dualblock, right=of p1] (d1) {$\Phi_\rmD^{1}\big(\,\bblambda_0, \bbx_0, \bbz\,;\theta_\rmD^{1}\big)$};
      \node[dualblock, below=of d1] (d2) {$\Phi_\rmD^{2}\big(\,\bblambda_1, \bbx_1, \bbz\,;\theta_\rmD^{2}\big)$};
      \node[vdots, below=of d2] (dd) {$\vdots$};
      \node[dualblock, below=of dd] (dL) {$\Phi_\rmD^{L}\big(\,\bblambda_{L-1}, \bbx_{L-1},\bbz\,;\theta_\rmD^{L}\big)$};
      \draw[arrow] (d1) -- (d2);
      \draw[arrow] (d2) -- (dd);
      \draw[arrow] (dd) -- (dL);
    
      \def\insetA{1.8cm}
      \def\insetB{1.1cm}
    
      \coordinate (busX_top) at ($(p1.north east)+(\insetA,-0.2)$);
      \coordinate (busX_bot) at ($(pL.south east)+(\insetA,-1.3)$);
      \coordinate (busL_top) at ($(p1.north east)+(\insetB,0.8)$);
      \coordinate (busL_bot) at ($(pL.south east)+(\insetB,-0.7)$);
    
      \draw[line width=2pt,teal] (busX_top) -- (busX_bot);
      \draw[line width=2pt,blue] (busL_top) -- (busL_bot);
    
      \coordinate (primal_ul) at ($(p1.north west)+(-0.6cm,1.0cm)$);
      \coordinate (primal_lr) at ($(pL.south east)+(0.6cm,-0.8cm)$);
      \draw[line width=1.0pt] (primal_ul) rectangle (primal_lr);
    
      \coordinate (x_mid) at ($(pL.south)+(0,-1.3cm)$);
      \draw[arrow] (pL.south) -- (x_mid) -- (busX_bot);
    
      \node[gray, above left=3pt and 0pt of p1.north] {$\widetilde\bbx_0$};
      \foreach \i/\lbx/\lbl in {1/1/0,2/2/1,L/K/L-1} {
        \coordinate (BX\i) at (busX_top |- p\i);
        \coordinate (BL\i) at (busL_top |- d\i);
        \coordinate (DW\i) at (d\i.west |- p\i);
        \coordinate (DS\i) at ($(d\i.north) + (0,0.1cm)$ |- p\i);
        \node[font=\small, above right=2pt and 2pt of p\i.east] {$\bblambda_l$};
        \node[gray, below left=3pt and 0pt of p\i.south] {$\widetilde\bbx_{\lbx}$};
    
        \draw[arrow] (BL\i) -- (p\i);
        \draw[thick,redarrow] ($(DS\i)+(0,0.15)$) -- ($(BL\i)+(0,0.7)$);
        \node[font=\small, above right=2pt and 2pt of d\i.north] {$\bblambda_{\lbl}$};
        \draw[thick,tealarrow] (BX\i) -- node[midway,above] {$\!\bbx_{\lbl}$} (DW\i);
    

    \coordinate (p1_down) at ($(p1.north)+(0,0.7cm)$);
    \draw[arrow] (p1_down) -- (p1.north);
    \coordinate (d1_down) at ($(d1.north)+(0,0.7cm)$);
    \draw[arrow] (d1_down) -- (d1.north);
    
    \coordinate (dL_down) at ($(dL.south)+(0,-1.7cm)$);
    \draw[arrow] (dL.south) -- (dL_down);
    
    \draw[thick,redarrow] ($(dL.south)+(0,-0.4)$) -- ($(busL_bot)+(0,0.3)$);
    \node[font=\small, below right=30pt and 2pt of dL.south] {$\bblambda_L$};
    
    \coordinate (xL) at ($(dL.west)+(0,-1.25cm)$);
    \draw[arrow] (p1_down) -- (p1.north);
    \draw[thick,tealarrow] ($(busX_bot)+(0,0.45)$) -- (xL) -- ($(xL)+(0,-0.8)$);
    \node[font=\small, below right=45pt and 2pt of dL.west] {$\bbx_L$};
    
      }
    
    \end{tikzpicture}
}   
    \caption{The structure of the unrolled primal GNN (left) and dual GNN (right). The forward pass alternates between: i) each dual layer $\Phi_\rmD^l$ outputs $\bblambda_l$, which is sent to the primal network $\Phi_\rmP$ to query for the Lagrangian minimizer $\bbx^*(\bblambda_l)$, and ii) the primal network performs its internal forward pass and returns $\bbx_l \approx \bbx^*(\bblambda_l)$. 
    }
    \label{fig:architecture}
\end{figure}
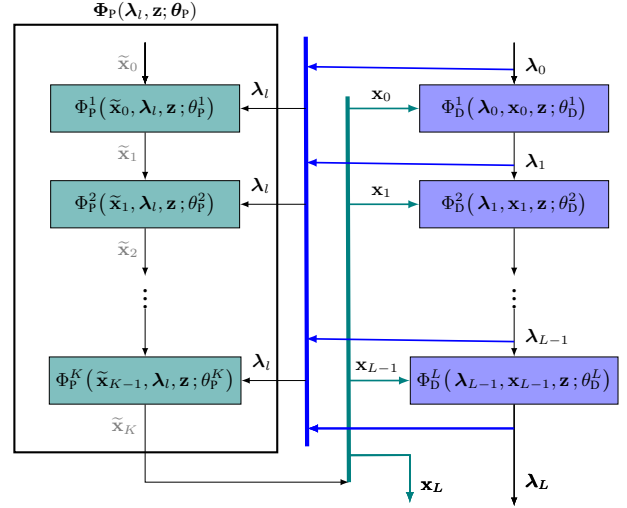

%% file: sections/training.tex
\section{Training with Descent Constraints} \label{sec:training}

We formulate the unsupervised training problem as a nested optimization problem that mirrors the dual problem. The outer problem is the dual network training problem over a distribution of problem instances, while the inner problem trains the primal network over a joint distribution of problem instances and multipliers. Formally, the nested training problem is defined as 
\begin{align}
     \bbtheta_\rmD^* ~\in~ \underset{\bbtheta_\rmD}{\text{argmax}} \quad &  \mbE_\bbz \Big[\ccalL \big(\, \bbPhi_\rmP (\bblambda_L, \bbz; \bbtheta_\rmP^*)  , \, \bblambda_L; \bbz\big) \Big], \label{eq:dual_training} \\
    {\text{with} \quad \,} &
        {\bbtheta_\rmP^*} \in \underset{\bbtheta_\rmP}{\text{argmin}} \ \mbE_{\bblambda,\bbz} \Big[\ccalL \big( \, \bbPhi_\rmP (\bblambda, \bbz; \bbtheta_\rmP), \, \bblambda; \bbz \big) \Big], \label{eq:primal_training}
\end{align}
where $\bblambda_L = \bbPhi_\rmD(\bbz;\bbtheta_\rmD, \bbtheta_\rmP^*)$ is the final output of the dual network. 
The joint training objective in \eqref{eq:dual_training}--\eqref{eq:primal_training} guides the final outputs but does not guarantee that the intermediate layer trajectories are monotonically descending (primal) or ascending (dual). 

To address this challenge, we impose descent constraints on the primal network such that the gradient norm of the Lagrangian with respect to $\bbx$ is forced to decrease, on average, across the layers. The primal network training is then cast as
\begin{align}
     \bbtheta_\rmP^* \in \underset{\bbtheta_\rmP}{\text{argmin}} \, & \mbE \Big[\,\ccalL \big( \, \bbPhi_\rmP (\bblambda, \bbz; \bbtheta_\rmP), \, \bblambda; \bbz \big) \,\Big], \nonumber \\
    {\text{s.t. \, }} &
        \mbE \Big[ {
        \|\!\nabla_\bbx\ccalL\big(\widetilde\bbx_k, \bblambda; \bbz\big)\!\| \! - \! \alpha_k 
        \|\!\nabla_\bbx \ccalL\big(\widetilde\bbx_{k-1}, \bblambda; \bbz\big)\!\|} \Big] \!\leq\! 0, \forall k\label{eq:constrained_primal_training}
\end{align}
where $\alpha_k$ is a design parameter that controls the descent rate and $\widetilde\bbx_0$ is initialized randomly.
Analogously, we impose ascent constraints on the dual network to ensure that the gradient norm of the Lagrangian with respect to $\bblambda$, i.e., $\|\bbf(\cdot; \bbz)\|$, decreases across the layers. This is a proxy for decreasing the gradient of the dual function (Danskin's theorem \cite{bertsekas1997nonlinear}).
The dual network training is then defined as 
\begin{align}
     \bbtheta_\rmD^* ~\in~ \underset{\bbtheta_\rmD}{\text{argmax}} \ \  & \mbE \Big[\, \ccalL \big(\, \bbPhi_\rmP (\bblambda_L, \bbz; \bbtheta_\rmP^*)  , \, \bblambda_L; \bbz\big) \,\Big], \nonumber \\
    {\text{s.t. \quad}} &
        \mbE\Big[\,\big\| \bbf(\bbx_{l}; \bbz) \big\| - \beta_l  \big\|\bbf(\bbx_{l-1} ; \bbz) \big\|\,\Big] \leq 0, \
    \forall l, \label{eq:constrained_dual_training}
\end{align}
where $\beta_l$ is a design parameter and $\bblambda_0$ is randomly initialized. Problems \eqref{eq:constrained_primal_training} and \eqref{eq:constrained_dual_training} are constrained nonconvex programs with small duality gap \cite{chamon2022constrained}. Therefore, they can be tackled in the dual domain with meta dual variables, $\bbmu$ and $\bbnu$, via Algorithms \ref{alg:primal-training} and \ref{alg:dual-training}, respectively, in which the hat notation denotes the empirical averages.

In principle, the primal network should be trained first on the joint problem-multiplier distribution before training the dual network with the primal parameters held fixed. However, the multiplier distribution is generally unknown \emph{a priori} as it depends on the multipliers generated by the dual network during execution. The nested formulation in \eqref{eq:dual_training}--\eqref{eq:primal_training} provides a solution to this challenge by enabling an alternating training scheme. First, we freeze the primal parameters and train the dual network on the outer objective for a fixed number of epochs, producing layerwise multiplier trajectories. These trajectories then serve as training data for the primal network, which we update for a fixed number of epochs. We keep alternating between training the two networks to ensure that each is optimized with data generated by the current state of the other.  

Note that although we replace the original convex problem with nonconvex training problems solved with the same algorithm we unroll, the computational cost of the iterative algorithm is now paid offline. The iterative algorithm is used once during training, after which each new instance is solved by a single forward pass of a neural network, significantly lowering per-instance latency and compute.

\begin{algorithm}[t]
\caption{Primal Network Training}
\label{alg:primal-training}
\begin{algorithmic}[1]
\State Inputs: $\bbtheta_\rmP, \bbtheta_\rmD, \bbmu, \epsilon_\rmP, \eta_\rmP$
\For {each epoch}
    \For{each primal batch}
        \State Sample $\{\bbz_{(j)}\}_{j=1}^N \sim \ccalD_\bbz$ 
        %
        \State Sample $\{\bblambda_{(i,j)}\}_{i=1, j=1}^{M,N}$ from the trajectories by $\bbtheta_\rmD$
        \State Execute the primal network to generate $\{ \widetilde\bbx_{k,(i,j)} \}_{k,i,j}$ 
        %
        \State $\ell(\bbtheta_\rmP) \leftarrow \widehat\ccalL (\widetilde\bbx_{K}, \bblambda; \bbz )$ 
        %
        \State {$\ccalC_k(\bbtheta_\rmP, \bbmu) \!\leftarrow \! \mu_k 
        \Big(\!\|\! \widehat\nabla\ccalL(\widetilde\bbx_{k}, \bblambda;\bbz)\!\| \!- \!\alpha_k \!\|\! \widehat\nabla\ccalL(\widetilde\bbx_{k-1}, \bblambda; \bbz)\!\|\!\Big)$}
        %
        \State $\bbtheta_\rmP \leftarrow \bbtheta_\rmP - \epsilon_\rmP \cdot \Big(\nabla \ell(\bbtheta_\rmP) + \nabla_{\bbtheta_\rmP} \sum_k\ccalC_k(\bbtheta_\rmP, \bbmu)\Big)$ 
        %
        \State $\bbmu \leftarrow \big[\, \bbmu + \eta_\rmP \cdot \nabla_{\bbmu} \ccalC(\bbtheta_\rmP, \bbmu) \,\big]_+$ 
    \EndFor
\EndFor
\State \Return $\bbtheta_\rmP, \bbmu$
\end{algorithmic}
\end{algorithm}

%% file: sections/numerical.tex
\section{Numerical Results}\label{sec:simulations}

\begin{figure*}[t]
    \centering
    \vspace{-0.4cm}
    \includegraphics[width=\textwidth]{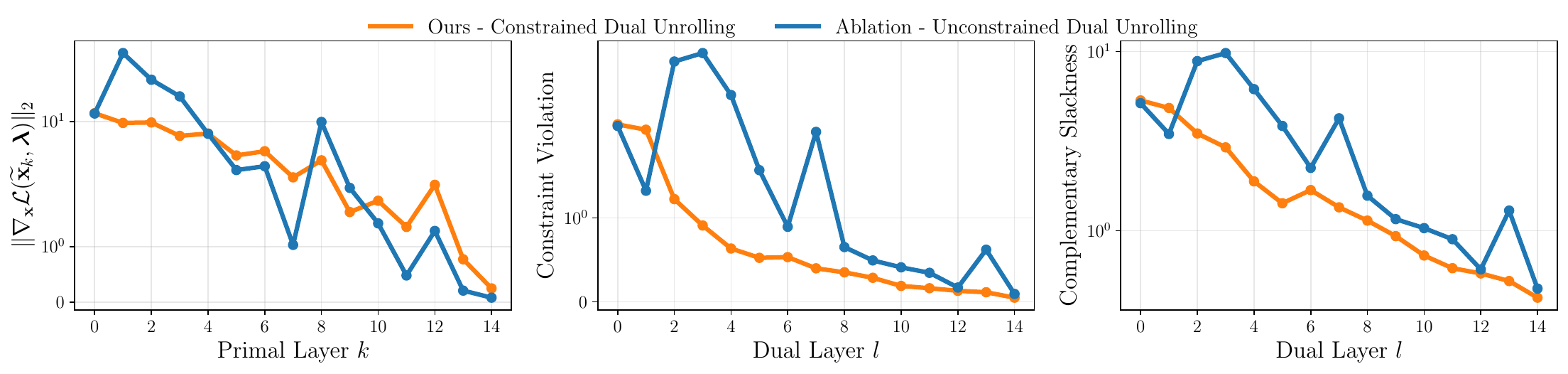}
    \caption{Descent Guarantees. (Left) The gradient norm of the Lagrangian across the primal layers. (Middle) The constraint violation across the dual layers. (Right) The complementary slackness $\bblambda_L^\top \bbf(\bbx_l)$ across the unrolled dual layers. 
    The constrained model exhibits a consistent decrease in all three quantities across layers, whereas the unconstrained model shows a more oscillatory pattern.
}
    \label{fig:miqp_ablation}
\end{figure*}
We consider a mixed-integer quadratic program (MIQP) with linear inequality constraints, formulated as:
\begin{align} 
        \min_{\bbx} \quad \tfrac{1}{2} \ & \bbx^\top \bbP \bbx + \bbq^\top \bbx \label{eq:miqp}\\
        \text{s.t.}\quad \quad \ \  & {\bar\bbA \bbx} \leq {\bar\bbb}, \label{eq:linear_constraints}\\
        \quad & {x}_i \in \{ -1, 1\}, \ \forall i \in {\cal I} \label{eq:integer_constraints},
\end{align}
where $\bbx \in \reals^n$, $\bbq \in \reals^n$, $\bar\bbA \in \reals^{m \times n}$, $\bar\bbb \in \reals^m$, $\bbP \in \reals^{n \times n}$ is PSD, and $\ccalI$ is a set that contains the indices of the binary variables with cardinality $|\ccalI|=r \leq n$.
Solving the MIQP in \eqref{eq:miqp}--\eqref{eq:integer_constraints} is NP-hard because of the binary constraints.
To mitigate this issue, we consider a convex relaxation of \eqref{eq:integer_constraints} in the form of box constraints, i.e., $-1 \leq x_i \leq 1, \ \forall i \in \ccalI$.
The linear constraints in \eqref{eq:linear_constraints} can be combined with the new ones, producing the relaxed MIQP problem,
\begin{align} 
         \min_{\bbx} \quad \tfrac{1}{2} \ & \bbx^\top \bbP \bbx + \bbq^\top \bbx \quad \text{s.t} \quad{\bbA \bbx} \leq {\bbb}, \label{eq:relaxed_miqp}
\end{align}
where $\bbA  \in \reals^{(m+2r) \times n}$ and $\bbb \in \reals^{m+2r}$. 
Specifically, we define $\bbA = \big[ \bar\bbA; \, \bbM; \, -\bbM   \big]$ and $\bbb = \big[ \bar\bbb \, ; \mathbf{1}_r \,; \mathbf{1}_r \big]$. Here, $\bbM \in \{0, 1\}^{r \times n}$ is a selection matrix whose $j$-th row is the standard basis vector $\bbe_{i_j}^\top$ for $i_j \in \ccalI$,
and $\mathbf{1}_r$ is an $r$-dimensional all-ones vector.
The Lagrangian function associated with the relaxed problem \eqref{eq:relaxed_miqp} is defined as
$
    \ccalL(\bbx, \bblambda) = \tfrac{1}{2} \bbx^\top \bbP \bbx + \bbq^\top \bbx + \bblambda^\top \big( 
                                \bbA \bbx - \bbb
                                \big).
$

\begin{figure*}[t]
    \centering
    \vspace{-10pt}
    \includegraphics[width=\textwidth]{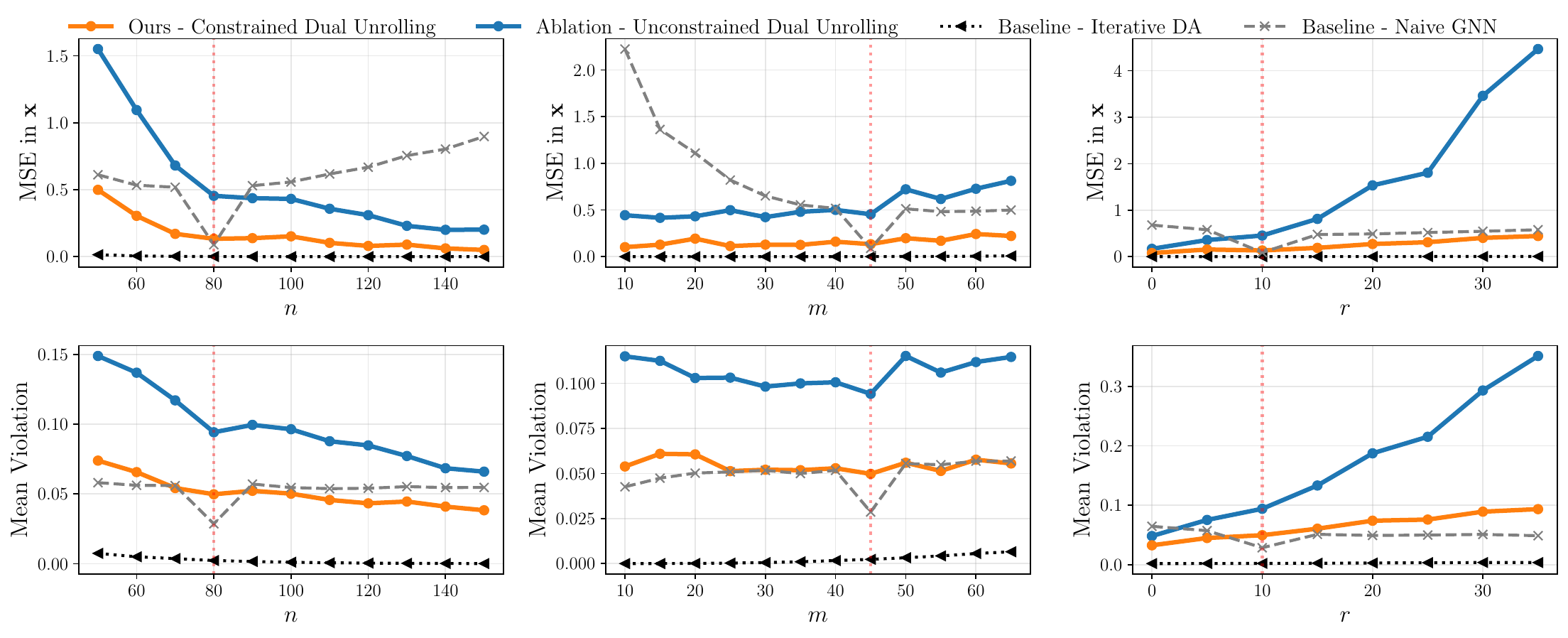}
    \caption{Robustness under OOD problems, varying (left) the number of optimization variables $n$, (middle) the number of linear constraints $m$, and (right) the number of integer-valued variables $r$. The red dotted line represents the in-distribution scenario: $n=80, m=45,$ and $r=10$. 
    Our constrained dual unrolling outperforms the other learning-based methods in optimality (top) and feasibility (bottom) across all OOD scenarios.
    }
    \label{fig:miqp_ood}
    \vspace{-10pt}
\end{figure*}

\begin{algorithm}[t]
\caption{Dual Network Training}
\label{alg:dual-training}
\begin{algorithmic}[1]
\State Inputs: $\bbtheta_\rmP,\bbtheta_\rmD, \bbnu, \epsilon_\rmD, \eta_\rmD$
\For {each epoch}
    \For {each dual batch}
        \State Sample $\{ \bbz_{(i)}\}_{i=1}^N \sim \ccalD_{\bbz}$
        %
        \State Execute the networks to generate $\{(\bbx_{l,(i)}, \bblambda_{l, (i)}) \}_{l,i}$
        %
        \State $\ell(\bbtheta_\rmD) \leftarrow - \widehat\ccalL \big(\bbx_{L} , \bblambda_{L}; \bbz\big) $
        %
        \State $\ccalC(\bbtheta_\rmD, \bbnu) \leftarrow
        \sum_l \nu_l \cdot
        \widehat\mbE\Big[\, \big\| \bbf(\bbx_{l}; \bbz \big\| - \beta_l  \big\|\bbf(\bbx_{l-1} ; \bbz) \big\| \,\Big]$
        %
        \State $\bbtheta_\rmD \leftarrow \bbtheta_\rmD - \epsilon_\rmD \cdot \Big(\nabla \ell(\bbtheta_\rmD) + \nabla_{\bbtheta_\rmD} \ccalC(\bbtheta_\rmD, \bbnu)\Big)$
        %
        \State $\bbnu \leftarrow \big[\, \bbnu + \eta_\rmD \cdot \nabla_{\bbnu} \ccalC(\bbtheta_\rmD, \bbnu) \,\big]_+$  
    \EndFor
\EndFor
\State \Return $\bbtheta_\rmD, \bbnu$
\end{algorithmic}
\end{algorithm}

To facilitate using primal and dual GNNs, we model the MIQP problem with the graph adjacency, where the $n$ variable nodes preceding the $m+2r$ constraint nodes for notational convenience,
\begin{align}
    {\bf S} ~=~ \left[
                    \begin{matrix}
                        {\bf P} & {\bf A}^\top \\
                        {\bf A} & {\bf 0}
                    \end{matrix} \right]. \label{eq:GSO}
\end{align}
In our models, the $\ell$-th unrolled layer consists of a cascade of $T$ graph convolutional sub-layers \cite{gama_graphs_2020}. The $t$-th sub-layer filters the previous output $\bbX_{t-1}^{(\ell)}$ and maps it to 
\begin{align}
\bbX_{t}^{(\ell)} ~=~  \varphi  \left( \,  \sum_{h = 0}^{K_\rmh} \, 
                                    \bbS^h \bbX_{t-1}^{(\ell)} \bbTheta_{t,h}^{(\ell)}  \right), 
                                    \label{eq:graph_layer}
\end{align}
where $\bbTheta_{t,h}^{(\ell)} \in \reals^{F_{t-1} \times F_{t}}$ is the set of learnable parameters, $K_\rmh$ represents the filter taps, and $\varphi$ is a nonlinear activation function.

In the primal network, the input to the $k$th unrolled layer is
\begin{align}
    \widetilde\bbX_0^{(k)} ~=~ \left[
    \begin{matrix}
        \widetilde\bbx_{k-1} & \bbq \\
        \bblambda & \bbb
    \end{matrix} \right],
\end{align}
where $\widetilde\bbx_{k-1}$ is the output of the previous unrolled layer, and $\bblambda$, $\bbq$ and $\bbb$ are input data.
The output of the unrolled layer is then
\begin{equation}
    \widetilde\bbx_k ~=~ \widetilde\bbx_{k-1} +  \bbM_\rmP \, \widetilde\bbX_{T}^{(k)} \,  \bbW_k + \bbc_k ,
\end{equation}
where $\widetilde\bbX_{T}^{(k)}$ is the output of the $T$-th graph sub-layer, and $\bbW_k \in \reals^{F_{T}}$ and $\bbc_k \in \reals^{n}$ are the parameters of the readout layer. The selection matrix $\bbM_\rmP$ extracts the outputs associated with the $n$ variable nodes.
Similarly, the input to each unrolled dual layer is constructed as 
\vspace{-1pt}
\begin{align}
    \bbX_0^{(l)} ~=~ \left[
    \begin{matrix}
        \bbx_{l-1} & \bbq \\
        \bblambda_{l-1} & \bbb
    \end{matrix} \right],
\end{align}
\vspace{-1pt}
where $\bblambda_{l-1}$ is the previous dual estimate and $\bbx_{l-1}$ is the corresponding estimate of the primal network.
The output of the unrolled layer is expressed as
\vspace{-2pt}
\begin{equation}
    \bblambda_l ~=~ \varphi_\text{relu} \Big( \bby_{l-1} + \bbM_\rmD \, \bbX_{T}^{(l)} \, \bbW_l + \bbc_l \Big),
\end{equation}
\vspace{-2pt}
where $\bbM_\rmD$ selects the constraint-node values, and $\bbW_l \in \reals^{F_{T}}$ and $\bbc_l \in \reals^{m+2r}$ are learnable parameters---distinct from those of the primal layers despite the shared notation.

\textbf{Experiment Setup.}
We construct a dataset of $800$ MIQP problems with $n = 80$, $m=45$ and $r = 10$. 
Both primal and dual networks consist of $K = 14$ unrolled layers with $T = 3$ graph sub-layers. In both networks, each graph filter aggregates information from $K_\rmh = 1$ hop neighbors, processes $F_l = F_k = 32$ hidden features, and is followed by a tanh activation function.
After a grid search, we set the learning rates to: $\epsilon_\rmP = 10^{-4}$, $\epsilon_\rmD = 7 \times 10^{-4}$, $\eta_\rmP = 10^{-4}$ and $\eta_\rmD = 10^{-3}$. The constraint parameters $\alpha_k = 0.98$ and $\beta_l = 0.95$ are shared across the layers. For evaluation, we compare our approach to its unconstrained counterpart, which resembles the one in \cite{Park_Van_Hentenryck_2023}, except that we use unsupervised training losses rather than their supervised objective. We also compare to a standard GNN trained with a supervised loss \cite{chen2024expressive} to predict the primal solutions.

\textbf{Performance.} We evaluate the trained primal and dual GNNs over a testset of unseen $400$ problems. Fig. \ref{fig:miqp_ablation} shows that our models satisfy the descent constraints on the testset as they exhibit a consistent decrease in the Lagrangian gradient norm and in the mean constraint violations across the layers. In contrast, the ablated model, trained without descent constraints, exhibits more pronounced oscillations. These differences are not aesthetic, as the unconstrained model does not reach the same performance as the constrained one in predicting the true $(\bbx^*, \bblambda^*)$ pairs. This gap is more evident in Fig.~\ref{fig:miqp_ood} (the red dotted lines), where the constrained model attains a mean-squared error (MSE) of $0.133$ in $\bbx$ and a mean constraint violation of $0.049$, compared to $0.454$ and $0.094$ for the unconstrained model. This improvement comes with negligible overhead, since our formulation introduces one meta dual variable per layer, which does not alter the dominant computational components of the update.

Fig. \ref{fig:miqp_ood} also shows OOD results obtained by varying one problem parameter while keeping the others fixed. The distribution shift intensifies as $(m+2r)/n$ increases, yielding more challenging instances.
Across all OOD scenarios, our constrained model consistently outperforms its unconstrained counterpart in optimality and feasibility. The performance gap widens as the constraint-to-variable ratio $(m+2r)/n$ increases, suggesting that the descent constraints confer OOD robustness. 
The constrained model also surpasses the standard GNN, which is attributed to our unsupervised training objectives and to jointly training a dual network that predicts the optimal dual variable along with the primal solution. 
When the ODD multiplier distribution remains close to that of the in-distribution case, our model can produce solutions that match--or even exceed--in-distribution performance, whereas the naive GNN does not realize these gains.

%% file: sections/conclusions.tex
\section{Conclusions}
This paper introduced two coupled GNNs trained to jointly find the Lagrangian saddle point and forced to mimic the DA dynamics by imposing layerwise monotonic descent and ascent constraints. 
Once trained, solving new instances reduces to a single forward pass, significantly cutting per-instance latency and compute.
In MIQP experiments, the constrained models consistently outperformed both unconstrained unrolling and supervised GNN in OOD robustness (e.g., lower MSE and constraint violation).

\clearpage